\documentclass{sig-alternate-2013}

\usepackage{color}
\usepackage{tikz}
\usepackage{booktabs}
\usepackage{caption}
\usepackage{subcaption}
\usepackage{url}

\usepackage{color}

\usepackage{xspace}
\def\etal{\emph{et al.}\xspace}

\newfont{\mycrnotice}{ptmr8t at 7pt}
\newfont{\myconfname}{ptmri8t at 7pt}
\let\crnotice\mycrnotice%
\let\confname\myconfname%

\begin{document}
%

%
%
\title{The ImageNet Shuffle:\\Reorganized Pre-training for Video Event Detection}

\numberofauthors{1}
\author{
Pascal Mettes, Dennis C. Koelma, Cees G.~M. Snoek\\\\
\affaddr{University of Amsterdam}
}


\maketitle
\begin{abstract}
This paper strives for video event detection using a representation learned from deep convolutional neural networks. Different from the leading approaches, who all learn from the 1,000 classes defined in the ImageNet Large Scale Visual Recognition Challenge, we investigate how to leverage the complete ImageNet hierarchy for pre-training deep networks. To deal with the problems of over-specific classes and classes with few images, we introduce a bottom-up and top-down approach for reorganization of the ImageNet hierarchy based on all its 21,814 classes and more than 14 million images. Experiments on the TRECVID Multimedia Event Detection 2013 and 2015 datasets show that video representations derived from the layers of a deep neural network pre-trained with our reorganized hierarchy \textit{i)} improves over standard pre-training, \textit{ii)} is complementary among different reorganizations, \textit{iii)} maintains the benefits of fusion with other modalities, and \textit{iv)} leads to state-of-the-art event detection results. The reorganized hierarchies and their derived Caffe models are publicly available at
\url{http://tinyurl.com/imagenetshuffle}.
\end{abstract}

%
%

\section{Introduction}

The goal of this work is to detect events such as \emph{Renovating a home}, \emph{Birthday party}, and \emph{Attempting a bike trick} in web videos. The leading approaches~\cite{jiang2015fast,mettes2015bag,nagel2015event,xu2014discriminative} attack this challenging problem by learning video representations through a deep convolutional neural network~\cite{krizhevsky2012imagenet,szegedy2014going}. The deep network is pre-trained on a collection of 1,000 ImageNet classes~\cite{imagenet09,russakovsky2014imagenet}, used to  extract features for video frames, and then followed by a pooling operation over the frames to arrive at a video representation. We also learn representations for event detection with a deep convolutional neural network, but rather than relying on the default 1,000 class subset, we investigate how to leverage the complete ImageNet hierarchy for pre-training the representation. 

\begin{figure}[t]
\begin{subfigure}{\linewidth}
\includegraphics[width=\linewidth]{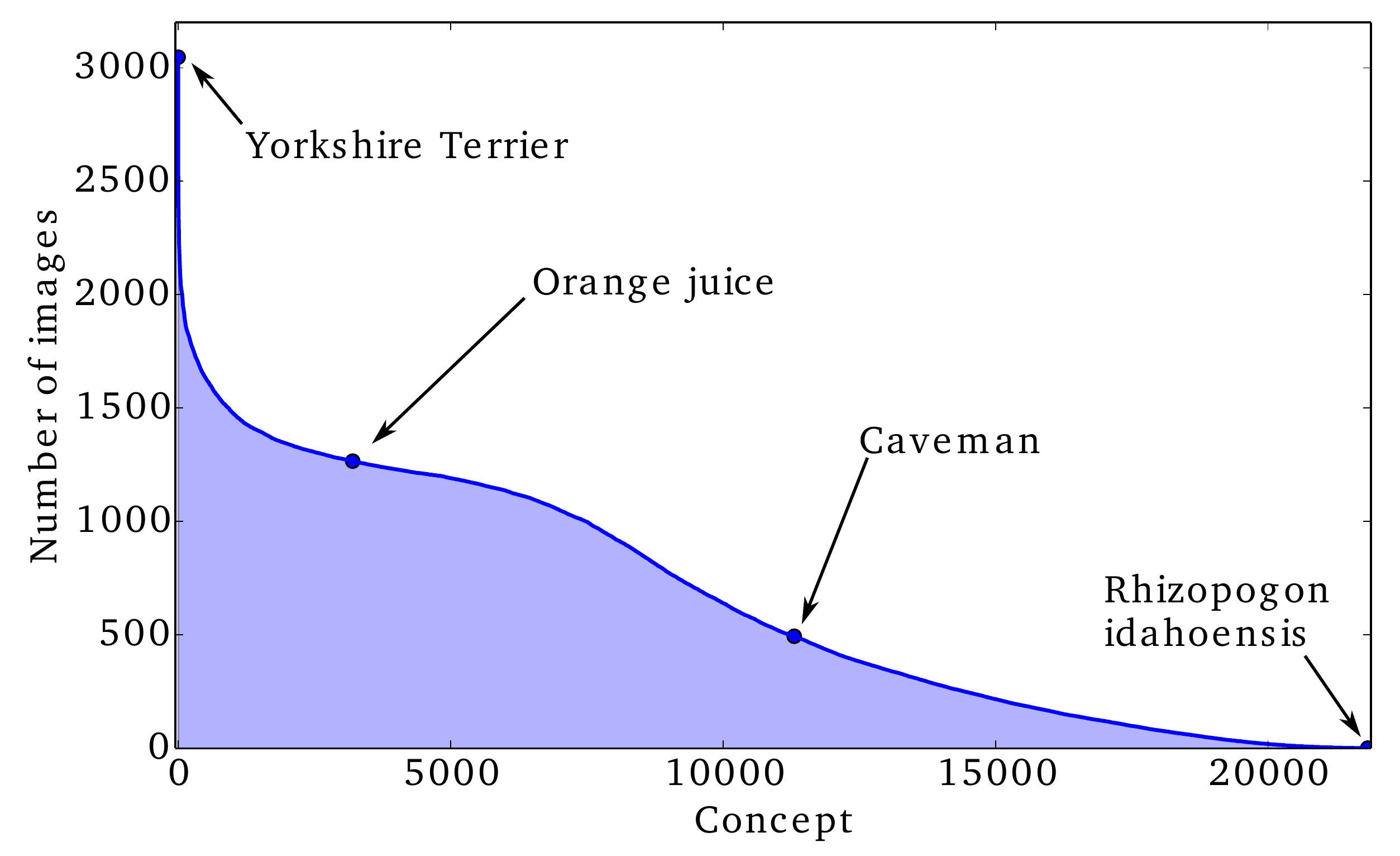}
\caption{Distribution of the number of images for the 21,814 ImageNet classes. Note the class imbalance.}
\label{fig:imbalance}
\end{subfigure}
\begin{subfigure}{\linewidth}
\centering
\includegraphics[width=0.49\linewidth]{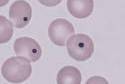}
\includegraphics[width=0.49\linewidth]{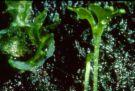}
\caption{An image of \emph{Siderocyte} (left) and \emph{Gametophyte} (right), two classes which seem over-specific for event detection.}
\label{fig:overspecific}
\end{subfigure}
\caption{Two problems when using the full ImageNet hierarchy for network pre-training: (a) image imbalance and (b) over-specific classes. In this work, we aim to reorganize the hierarchy into a balanced set of classes for more effective pre-training of video representations for event detection.}
\label{fig:problems}
\end{figure}

The complete ImageNet dataset consists of over 14 million images and 21,814 classes which are connected in a hierarchy as a subset of WordNet~\cite{miller1995wordnet}. State-of-the-art event detectors are pre-trained on a 1,000 class (1.2 million images) subset of ImageNet, as prescribed by the Large Scale Visual Recognition Challenge~\cite{russakovsky2014imagenet}. Hence, more than 90\% of the images in the hierarchy remain untouched during pre-training. We present an empirical investigation of the effect of using the full ImageNet dataset for event detection in web videos.

We identify two problems when trying to pre-train a deep network on the complete ImageNet hierarchy. First, there is an imbalance in the number of examples for each class, as is shown in Figure~\ref{fig:imbalance}. For example, the class \emph{Yorkshire Terrier} contains 3,072 images, whereas 296 other classes contain just a single image. Second, some classes seem over-specific for event detection in web videos. Consider for example the ImageNet categories \emph{Siderocyte} and \emph{Gametophyte} in Figure~\ref{fig:overspecific}. As a result, it seems suboptimal to directly pre-train a deep network on all 21,814 classes.

In this work, we introduce pre-training protocols that reorganize the full ImageNet hierarchy for more effective pre-training. The reorganization tackles the image imbalance and over-specific class problems. We propose two contrasting approaches that utilize the graph structure of ImageNet to combine and merge classes into balanced and reorganized hierarchies. We empirically evaluate our event detection using reorganized pre-training on the 2013 and 2015 NIST TRECVID Multimedia Event Detection datasets, for both datasets leading to state-of-the-art results. The Caffe models and detailed video feature extraction instructions are available at
\url{http://tinyurl.com/imagenetshuffle}.

\section{Related work}

\subsection{Event detection with pre-trained networks}

The state-of-the-art for event detection in videos focuses on video representations learned with the aid of deep convolutional neural networks~\cite{jiang2015fast,mettes2015bag,nagel2015event,xu2014discriminative}. The pipeline of these approaches consists roughly of three components. (1) A deep network is pre-trained on a large-scale image collection. Different deep networks have been employed for event detection, such as AlexNet~\cite{krizhevsky2012imagenet} in~\cite{jiang2015fast,mettes2015bag,nagel2015event} and VGGnet~\cite{simonyan2014very} in~\cite{xu2014discriminative}. (2)  Sampled video frames are fed to the network and features at fully-connected and/or soft-max layers are used as frame representations. (3) The frame representations are pooled into a fixed-sized video representation. A simple and effective pooling method is average pooling, where the frame representations are averaged over the video~\cite{karpathy2014large,mettes2015bag}. Recently, several works have shown that clustering deep frame representations into a codebook, followed by a VLAD~\cite{jegou2012aggregating} in~\cite{xu2014discriminative} or Fisher Vector encoding~\cite{sanchez2013image} in~\cite{nagel2015event}, leads to strong video representations. In this work, we aim for a similar pipeline of pre-training, frame representation, and video pooling. However, rather than relying on the standard pre-training protocol using 1,000 ImageNet classes, we leverage the complete ImageNet hierarchy for more effective pre-training.

Web videos provide a wide range of information about events, such as visual, motion, audio, and optical character information~\cite{over14}. Naturally, multiple works have investigated the fusion of information from different modalities~\cite{lan2014multimedia,myers2014evaluating,natarajan2012multimodal}. In this work, we also investigate the effect of fusing our deep representations with Motion Boundary Histogram (motion) features~\cite{wang13} and MFCC (audio) features~\cite{myers2014evaluating}, both of which are encoded into a video representation using Fisher Vectors~\cite{sanchez2013image}. This fusion allows us to compare the effectiveness of our deep representations to heterogeneous representations and to investigate how well our deep representations fare when combined with other sources of information.

\subsection{(Re)organizing hierarchies for events}

Various works have investigated the use of semantic hierarchies and ontologies for event detection. The work of Ye \etal~\cite{ye2015eventnet} focuses on hierarchical relations between events, to find a large collection of videos and event-specific concept classes. Their proposed EventNet has shown to yield an effective event detection~\cite{ye2015eventnet}. In our work, we focus on different hierarchical relations, namely between concept classes instead of events, to discover a general set of concepts for deep network pre-training. Other recent work on event detection has investigated relations among concept classes to rerank concept scores in the video representation~\cite{jiang2015fast}. We similarly focus on hierarchical relations among concept classes, but for the purpose of merging classes into a reorganized hierarchy for pre-training. For the hierarchy of ImageNet specifically, the work of Vreeswijk \etal~\cite{vreeswijk2012all} has shown that images from different layers of the hierarchy are visually different and that general concepts benefit from including linked concepts deeper in the hierarchy. We build upon these observations in our operations to reorganize the ImageNet hierarchy.

An alternative approach is to adjust concept hierarchies after feature extraction. For example, the selection of event-specific concepts based on the similarity to a textual event description has shown to yield effective event detection results without positive examples~\cite{jiang2015fast}. Mazloom \etal~\cite{mazloom2014conceptlets} show that concept selection is also beneficial for few-example event detection. Habibian \etal~\cite{habibian14} in turn, jointly learn a classifier for event detection and combine correlated concepts. Rather than changing the representations \emph{a posteriori} using text or video examples, we focus in this paper on reorganizing the hierarchical structure of visual ontologies \emph{before} event training,

\section{Reorganized pre-training}

\begin{figure}[t]
\includegraphics[width=\linewidth]{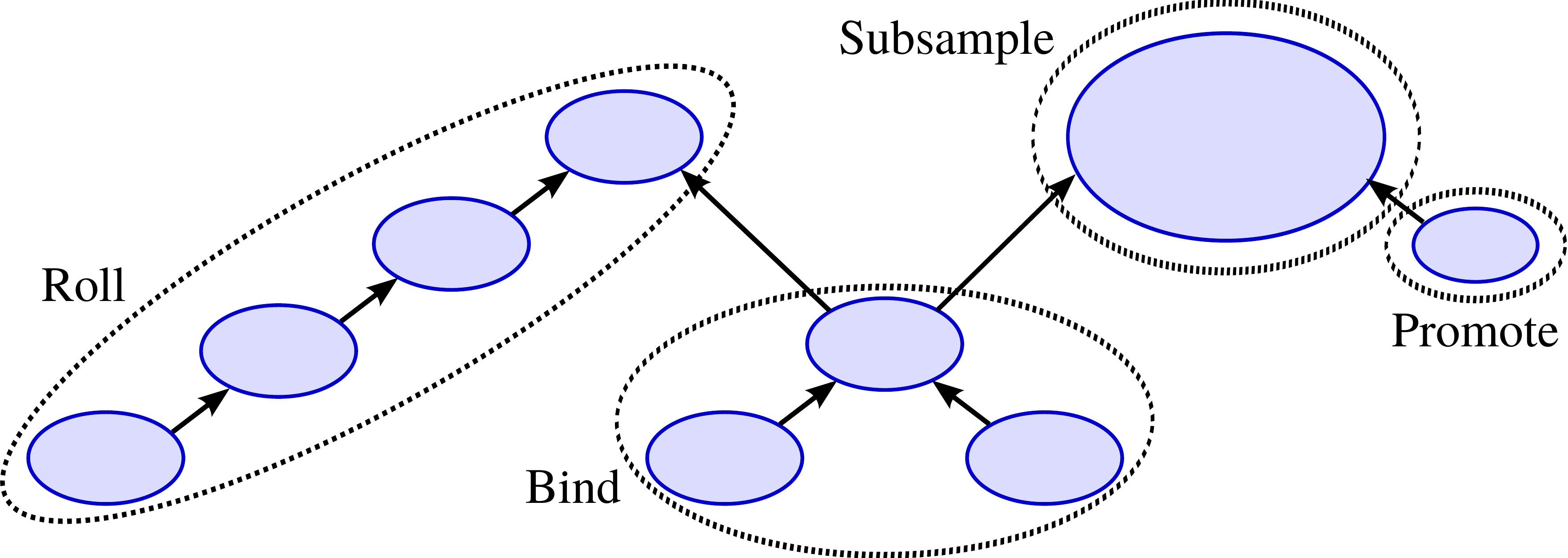}
\caption{Overview of where in the hierarchy the four operations are applied in the bottom-up approach. \textbf{Roll:} Roll up classes with a single chld-parent connection. \textbf{Bind:} Bind sub-trees for which the individual classes do not have enough images. \textbf{Promote:} Promote individual classes to their parent class if they do not have enough images. \textbf{Subsample:} Randomly subsample images from classes with too many images.}
\label{fig:reorganize}
\end{figure}

\begin{figure*}[t]
\centering
\begin{subfigure}{\textwidth}
\centering
\includegraphics[width=0.275\textwidth]{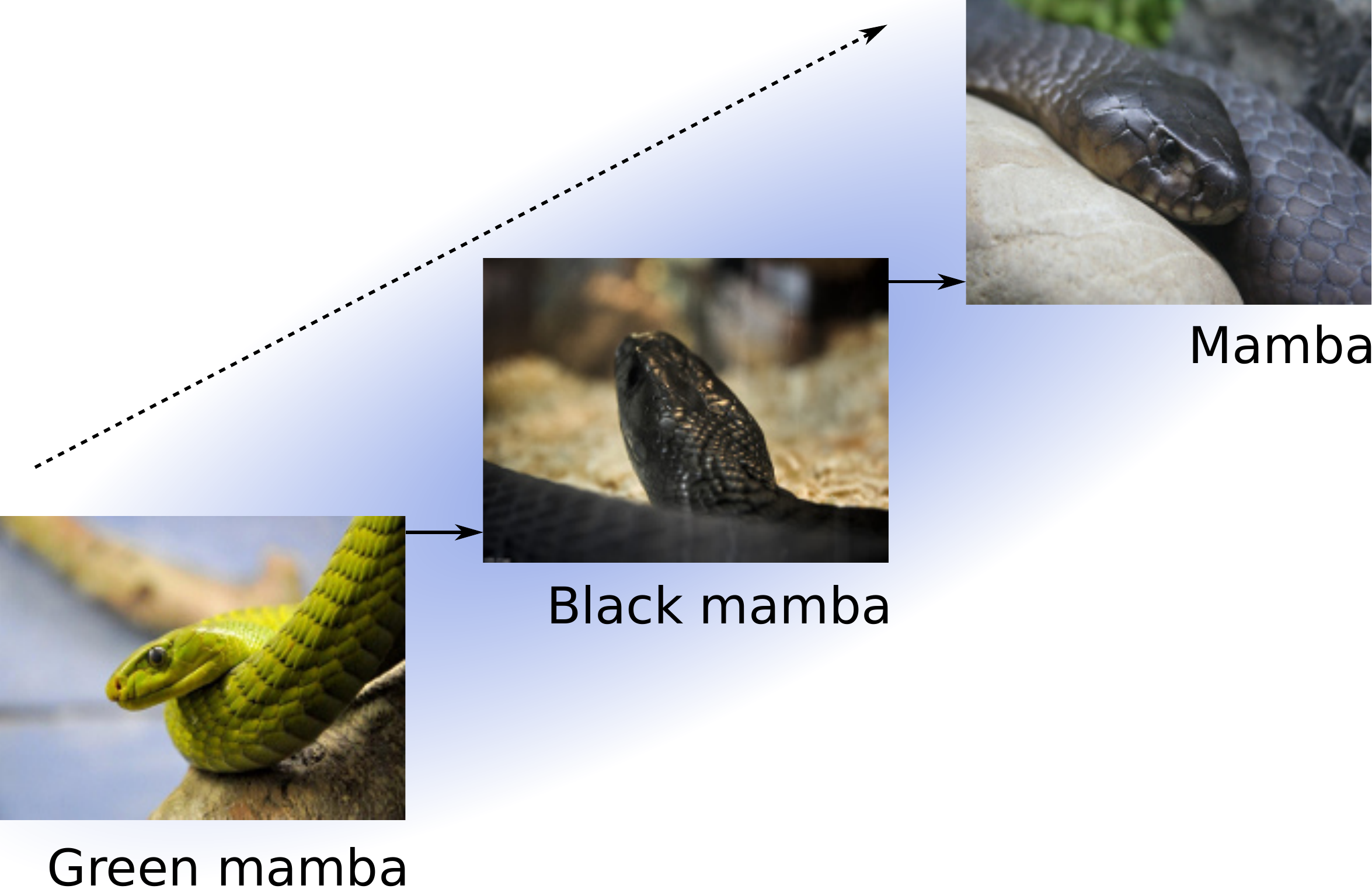}
\hspace{0.5cm}
\includegraphics[width=0.275\textwidth]{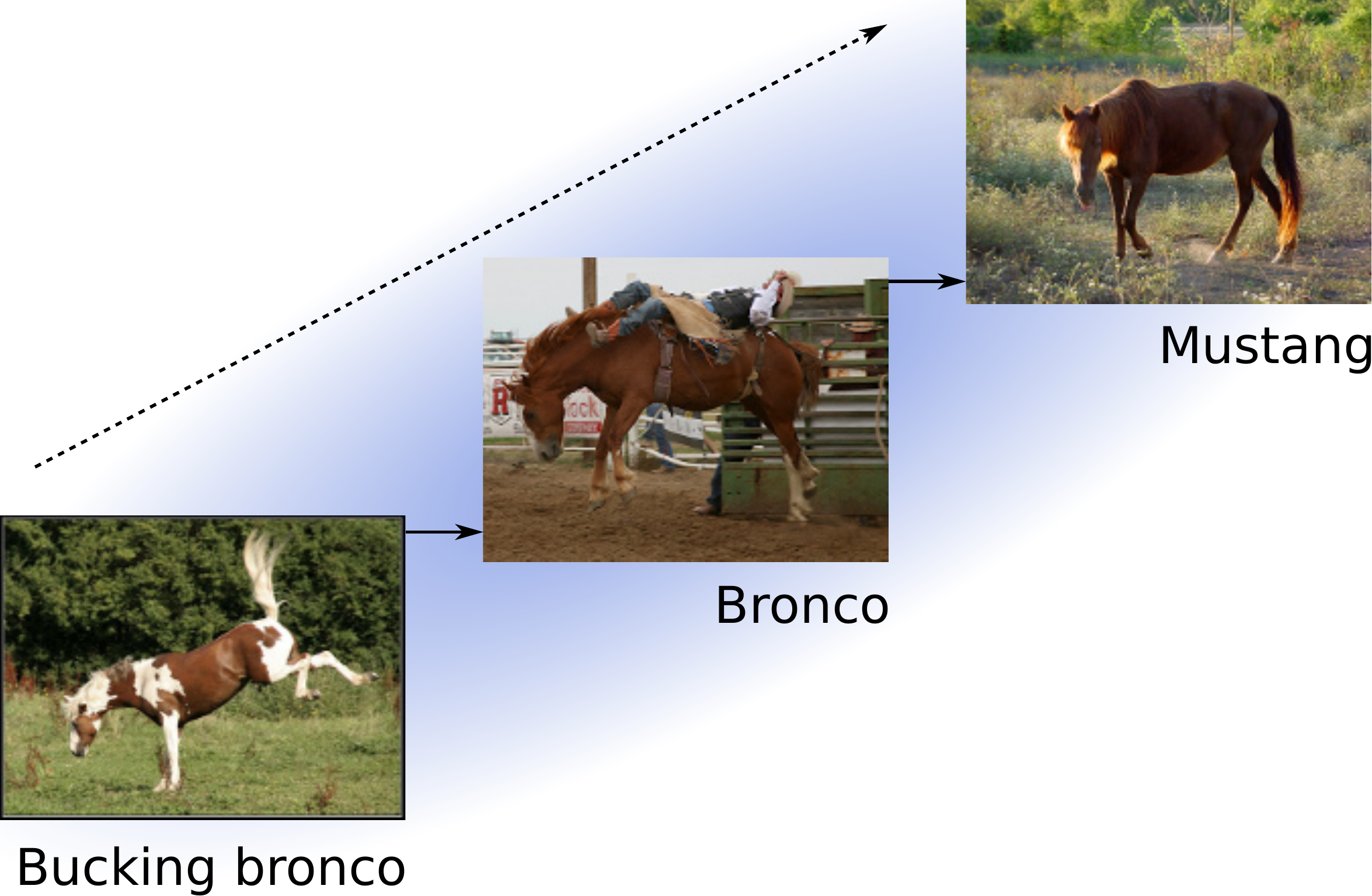}
\hspace{0.5cm}
\includegraphics[width=0.275\textwidth]{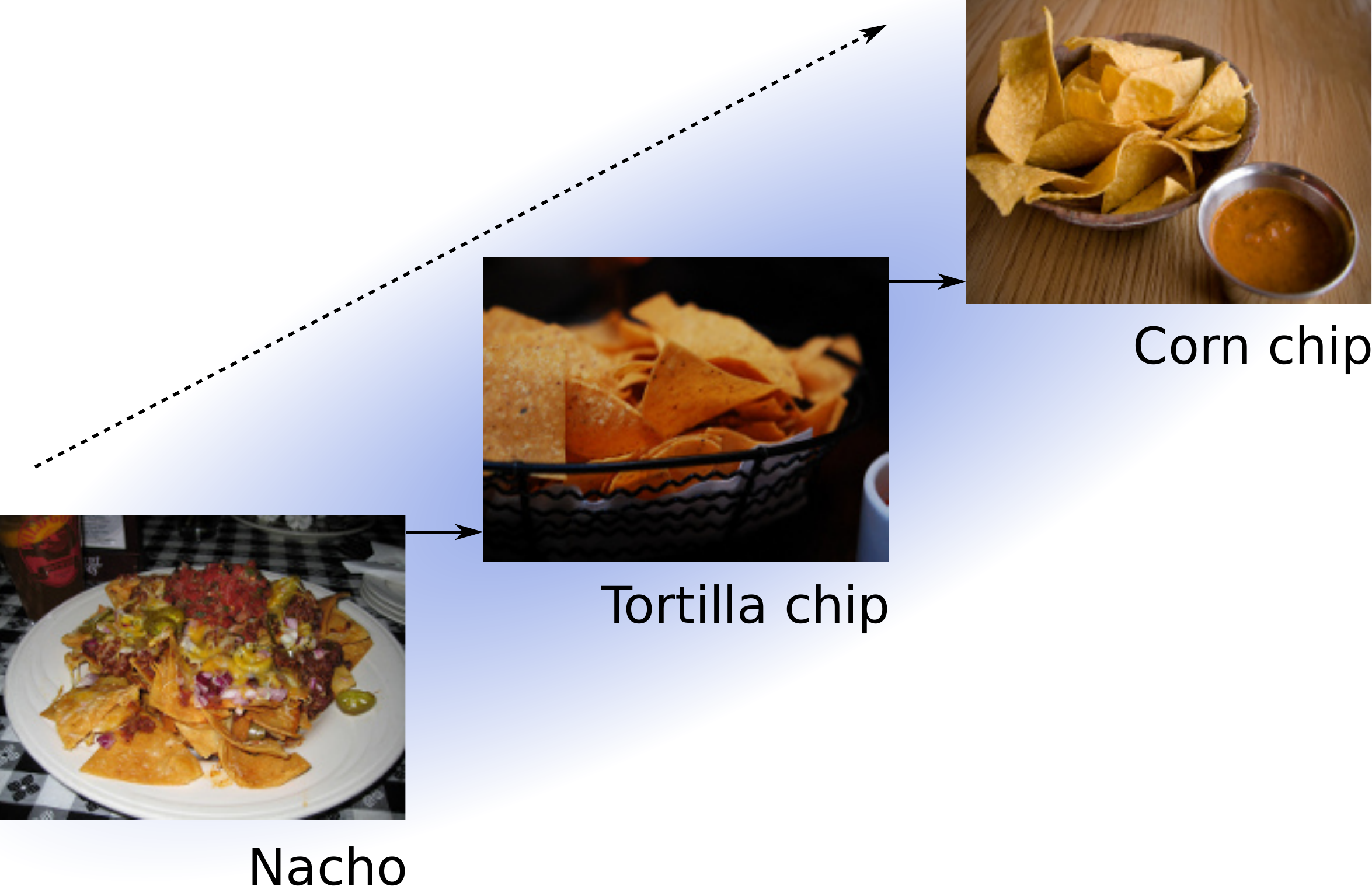}
\caption{Roll.}
\label{fig:roll}
\end{subfigure}
\noindent\rule[0.5ex]{\linewidth}{0.75pt}
\vspace{0.015cm}\\
\begin{subfigure}{\textwidth}
\centering
\includegraphics[width=0.27\textwidth]{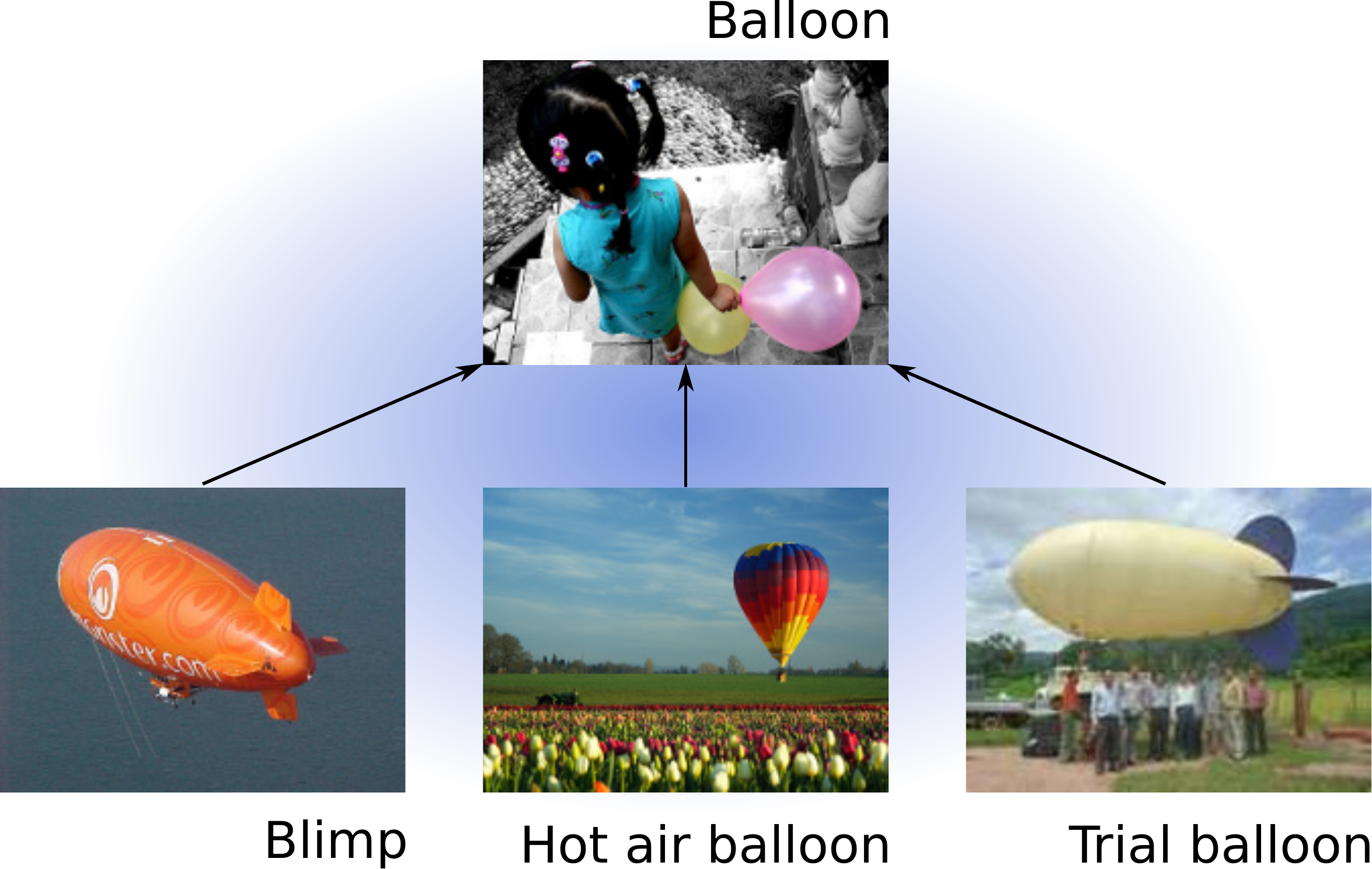}
\hspace{1.0cm}
\includegraphics[width=0.27\textwidth]{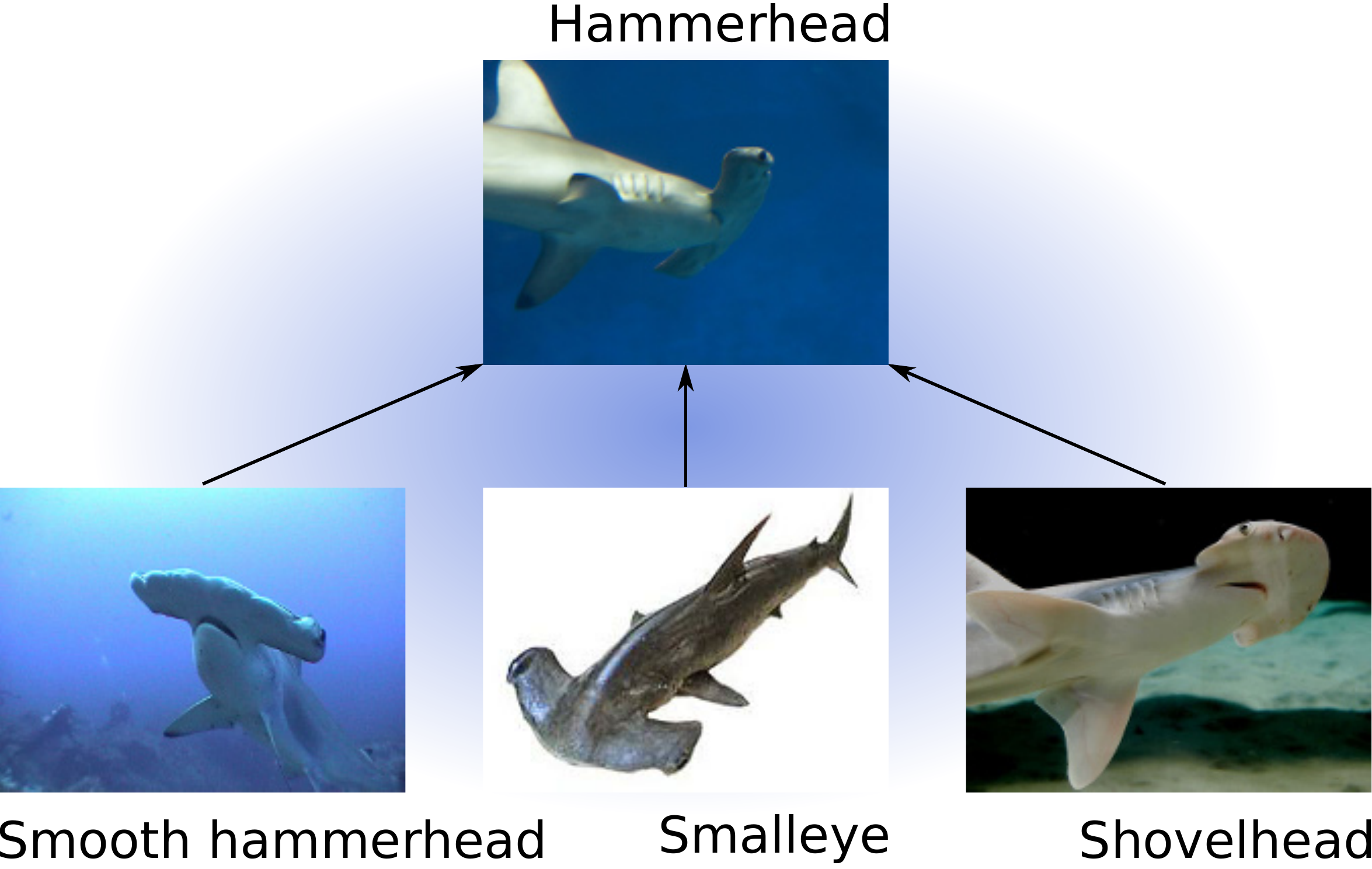}
\hspace{1.0cm}
\includegraphics[width=0.27\textwidth]{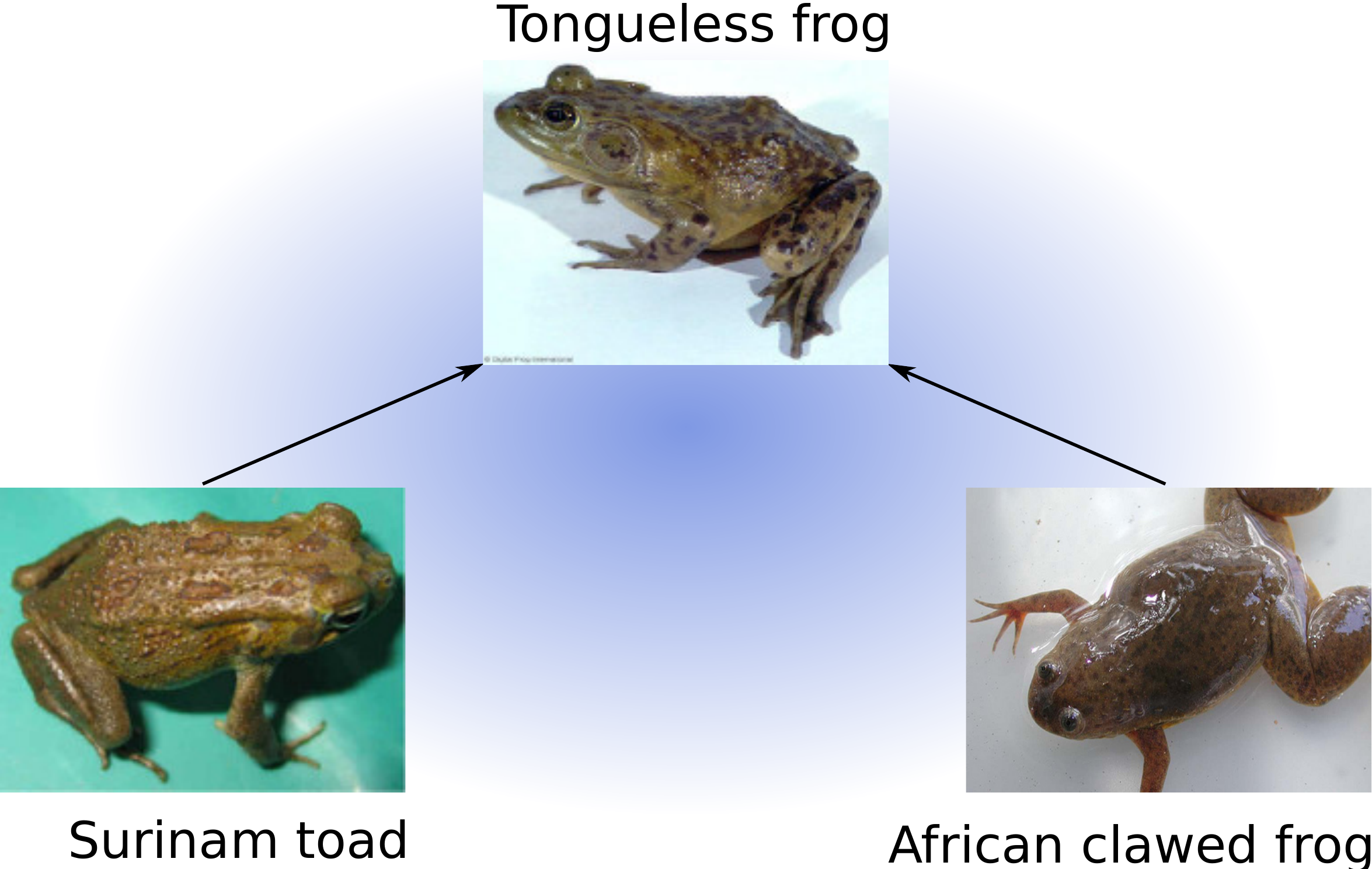}
\caption{Bind.}
\label{fig:bind}
\end{subfigure}
\noindent\rule[0.5ex]{\linewidth}{0.75pt}
\vspace{0.015cm}\\
\begin{subfigure}{\textwidth}
\centering
\includegraphics[width=0.2\textwidth]{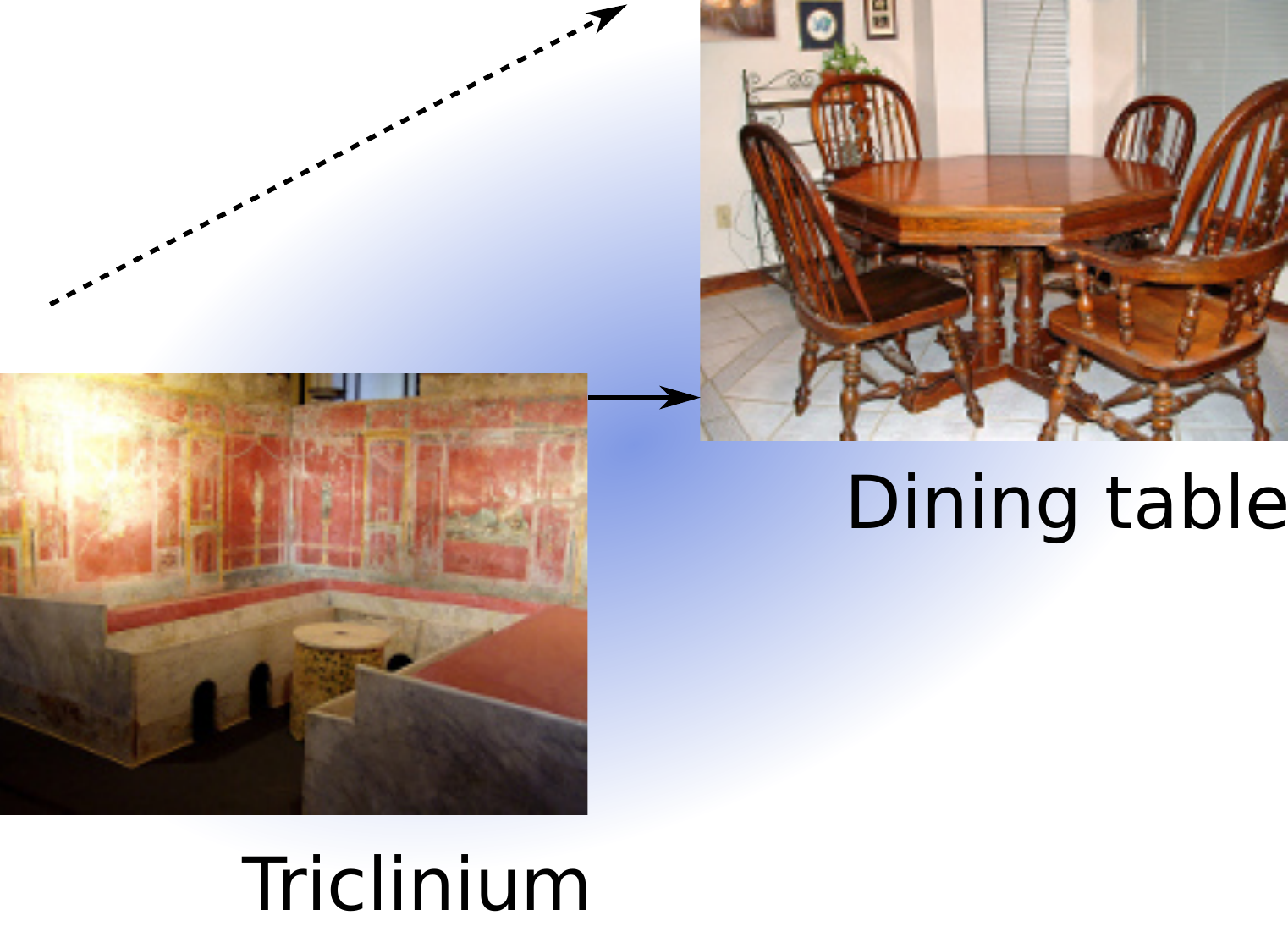}
\hspace{1.5cm}
\includegraphics[width=0.2\textwidth]{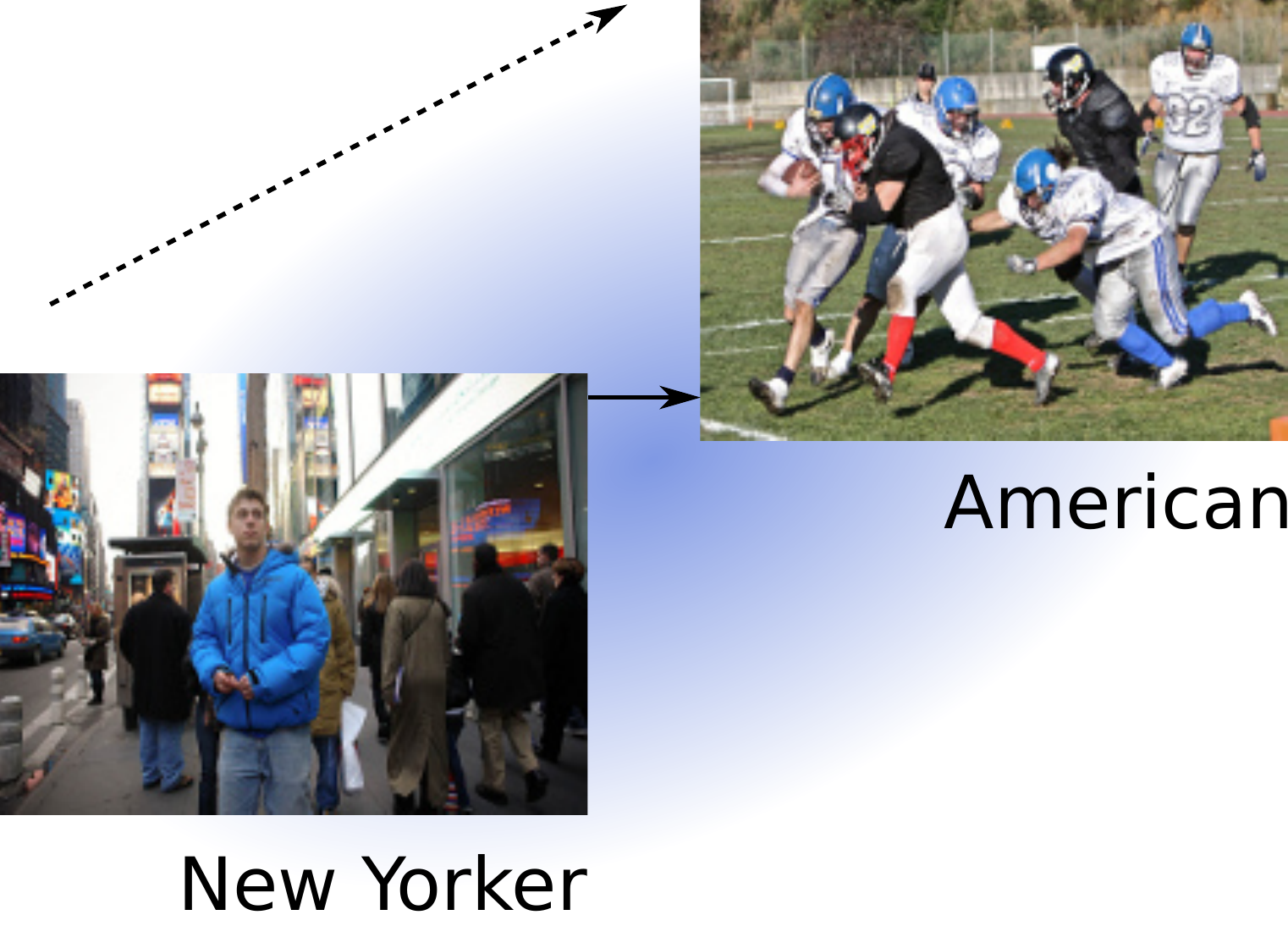}
\hspace{1.5cm}
\includegraphics[width=0.2\textwidth]{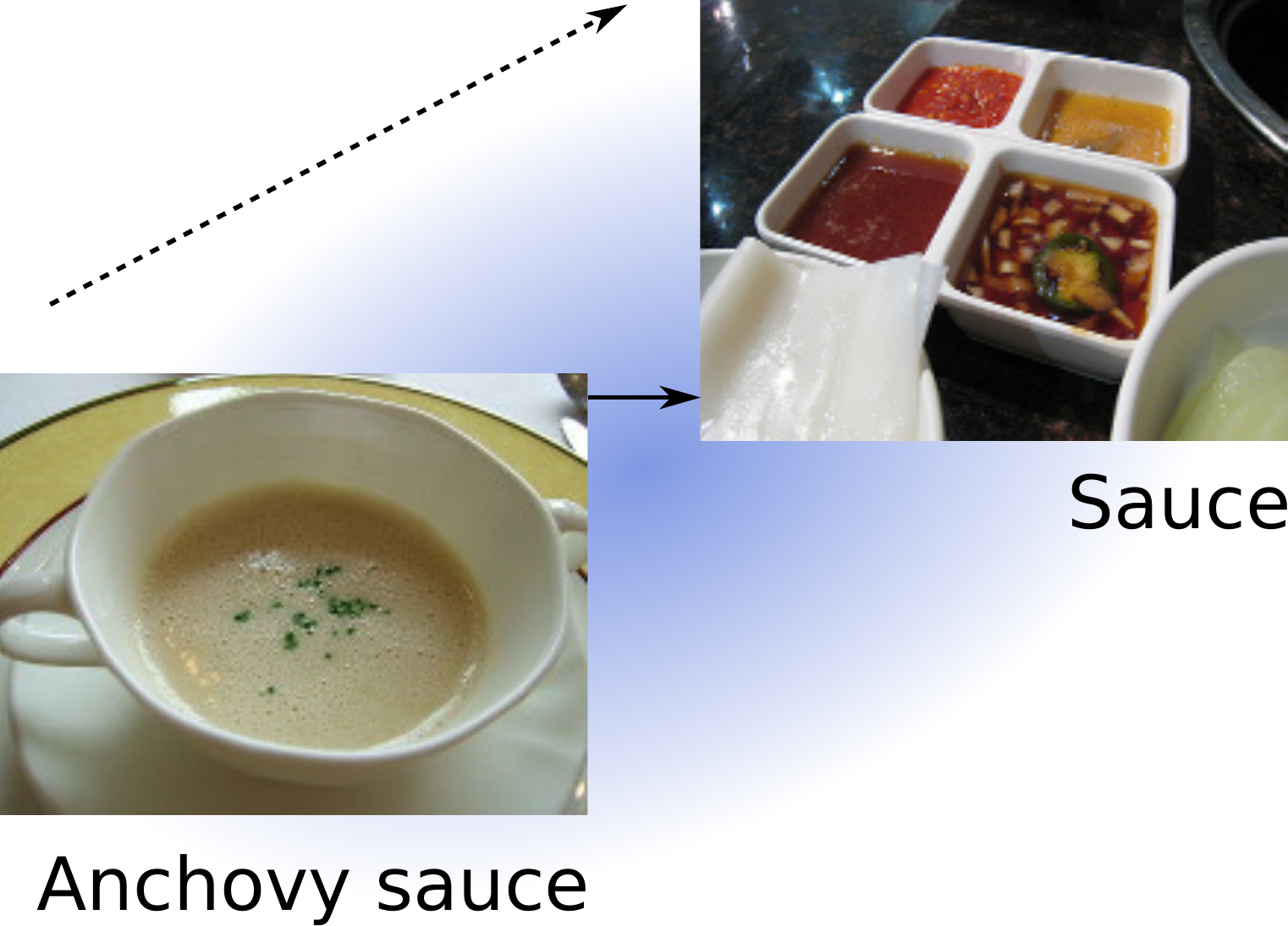}
\caption{Promote.}
\label{fig:promote}
\end{subfigure}
\noindent\rule[0.5ex]{\linewidth}{0.75pt}
\vspace{0.015cm}\\
\begin{subfigure}{\textwidth}
\centering
\includegraphics[width=0.745\textwidth]{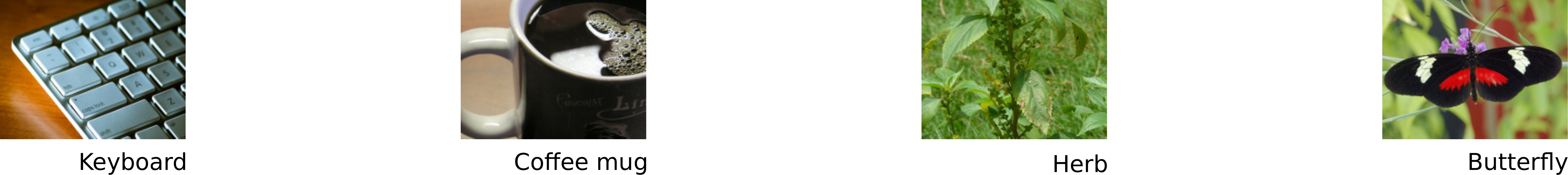}
\caption{Subsample.}
\label{fig:subsample}
\end{subfigure}
\caption{Visual examples for each of the four operations for the bottom-up reorganization.}
\label{fig:operations}
\end{figure*}

The classes in the ImageNet dataset are a subset of the WordNet collection~\cite{miller1995wordnet} and the classes are therefore connected in a hierarchy. The connectivity between classes provides information about their semantic relationship. We utilize the hierarchical relationship of WordNet for combining classes to generate reorganized ImageNet hierarchies for pre-training. We focus on two opposing approaches for reorganization, namely a bottom-up and top-down approach.

\subsection{Bottom-up reorganization}
For the bottom-up reorganization, we start from the original ImageNet hierarchy and introduce four reorganization operations. An overview of where in the hierarchy the four operations are performed is shown in Figure~\ref{fig:reorganize} and visual examples for each operation are shown in Figure~\ref{fig:operations}. We outline each operation separately.
\\
\textbf{Roll.} The roll operation is performed on single-link sub-trees of classes. In other words, the roll operation merges classes with a single child-parent connection, as shown in Figure~\ref{fig:reorganize} on the left. The motivation behind this operation is two-fold: \textit{i)} there is little semantic difference between a child and a parent if the parent has no other children. Treating the child and parent as separate classes during pre-training will dominate the backpropagation gradients to keep these classes separated. \textit{ii)} A single child of a parent is more likely to be over-specific for event detection. Single child-parent connections typically occur deep in the hierarchy, where details between classes become more fine-grained. In our evaluation, we indeed observe that the single child-parent connections occur predominantly in the deeper layers of the ImageNet hierarchy. Three chains of single child-parent connections are shown in Figure~\ref{fig:roll}. For example, the class \emph{Mamba} is a type of snake and has a single child, namely \emph{Black mamba}. In turn, the \emph{Black mamba} has a single child: \emph{Green mamba} (the green phase of the black mamba). In this example, we move all the images from the \emph{Black mamba} and \emph{Green mamba} classes to the \emph{Mamba} class.
\\
\textbf{Bind.} The bind operation is performed on sub-trees where the individual classes are sparse in the number of images. Let $S$ denote a sub-tree and let $c^i$ denote the number of images in class $c$. Then the bind operation is performed on sub-tree $S$ if $\sum_{c \in S} c^i < t_b$, where $t_b$ denotes the threshold on the number of images. The notion behind the bind operation is to deal with small and semantically coherent classes consisting of a parent and multiple children. The children individually do not contain enough images to treat them as separate classes. However, the combined set of parent and children forms a semantically consistent set with a desirable number of images. Three merged sub-trees that are combined with the bind operation are shown in Figure~\ref{fig:bind}. For example, the \emph{Hammerhead shark} has three children with a small number of images, namely \emph{Smooth hammerhead}, \emph{Smalleye hammerhead}, and \emph{Shovelhead}. Therefore, we opt to combine all these shark images into a single class.
\\\\
\textbf{Promote.} The promote operation is a unary operation. It is performed after the roll and bind have been performed. The promote operation simply promotes a class to its parent if its number of images is below a threshold $t_p$. This operation directly targets the imbalance problem, by adding images of classes with few examples to parent classes with more images. Figure~\ref{fig:promote} shows three cases of the promote operation. For example, the class \emph{Triclinium} (a dining table with couches at three sides) only contains 5 images. Therefore, the images are added to the \emph{Dining table} class, such that the \emph{Triclinium} images are still being used for pre-training without creating an imbalance in the hierarchy.
\\\\
\textbf{Subsample.} The subsample operation is also a unary operation and deals with the reverse problem of the other three operations. The subsample operation subsamples images from classes for which the number of images is above a threshold $t_s$. The reason for this operation is again for balancing purposes. If all images of over-populated classes are used in the stochastic optimization of the deep network, the network will overfit to these classes, resulting in suboptimal frame representations for event detection. Four examples of the subsample operation are shown in Figure~\ref{fig:subsample}, such as \emph{Keyboard}, \emph{Coffee mug}, and \emph{Herb}.
\\\\
We employ the defined operations in the described order. First, all single child-parent connections are rolled up. Second, all sub-trees in the hierarchy are binded based on threshold $t_b$ for their combined number of images. Third, all remaining classes with less than $t_p$ images are promoted to their parent. Fourth, during network pre-training, examples for all classes with more than $t_s$ images are randomly sub-sampled before the stochastic gradient descent optimization.


\subsection{Top-down reorganization}
An alternative and potentially complementary reorganization strategy is not to start from the deepest classes in the hierarchy, but from the head node. Here, we investigate a breath-first search approach. Let $t_t$ denotes a pre-specified threshold stating the minimum amount of images required for class in order to be used in the top-down reorganization. Then, starting from layer 0 in the hierarchy, i.e. the head node, we iteratively move down in the hierarchy and keep adding classes with at least $t_t$ images until we reach a desired amount of classes.

The breath-first search approach is outlined as follows. Let $l$ denote the previous layer of the hierarchy. We list all ImageNet classes in layer $l+1$ based on connections from classes in layer $l$ and order the classes in $l+1$ by their number of images. The sorting ensures that we select the classes with the highest number of images first, in case we reach the desired amount of classes before the end of the list. Then, we move through the ordered list and select all classes with at least $t_t$ images as long as the desired amount of selected classes is not reached. Afterwards, we move to the next layer and repeat the ordering and selection procedure.

By using a top-down approach, we ensure that only the most general classes are maintained for pre-training, while simultaneously keeping a balance in the image distribution through the threshold $t_t$.

\begin{figure*}[t]
\centering
\begin{subfigure}{0.49\textwidth}
\includegraphics[width=\textwidth]{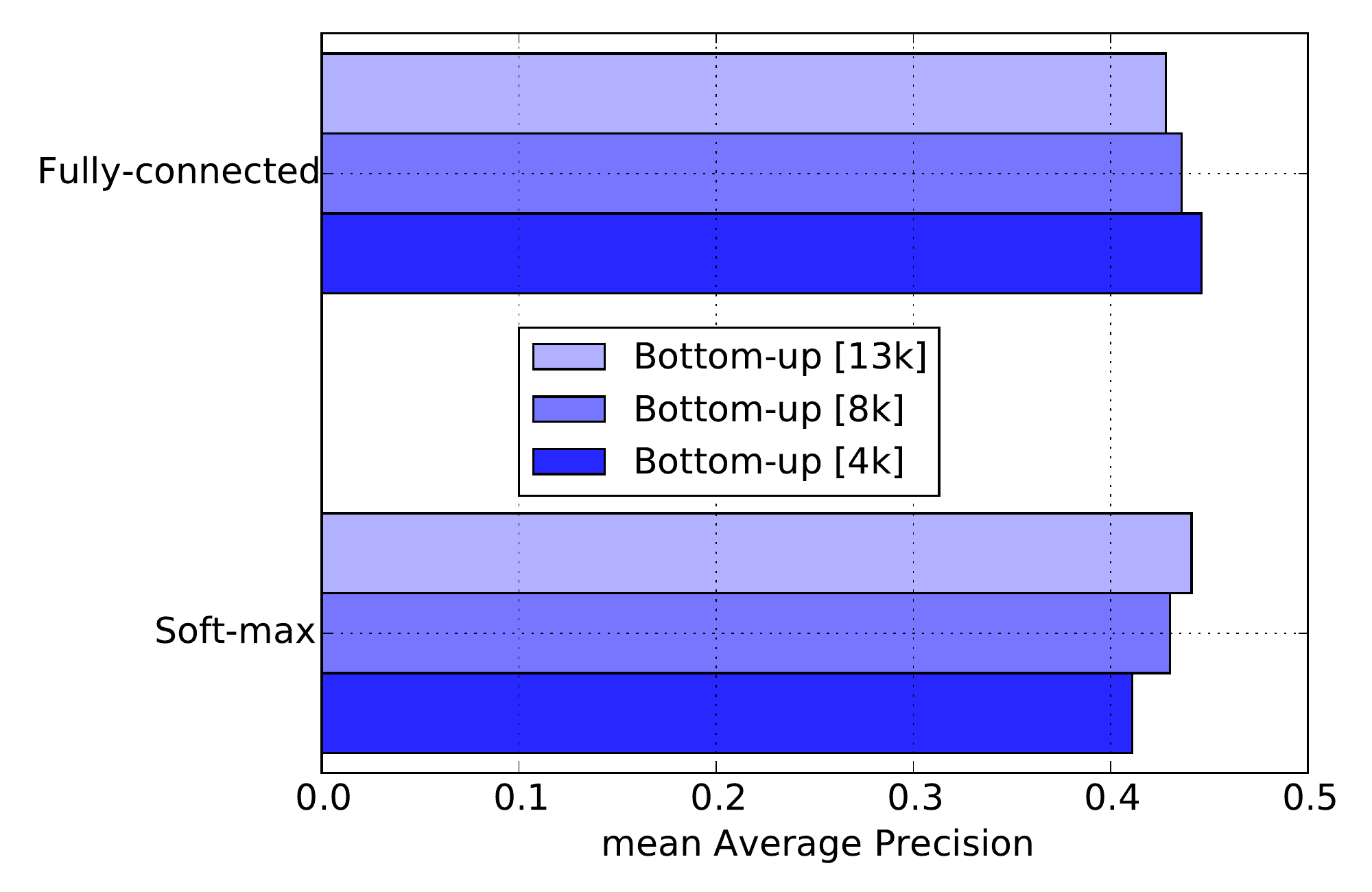}
\caption{TRECVID MED 2013 with 100 positives.}
\end{subfigure}
\begin{subfigure}{0.49\textwidth}
\includegraphics[width=\textwidth]{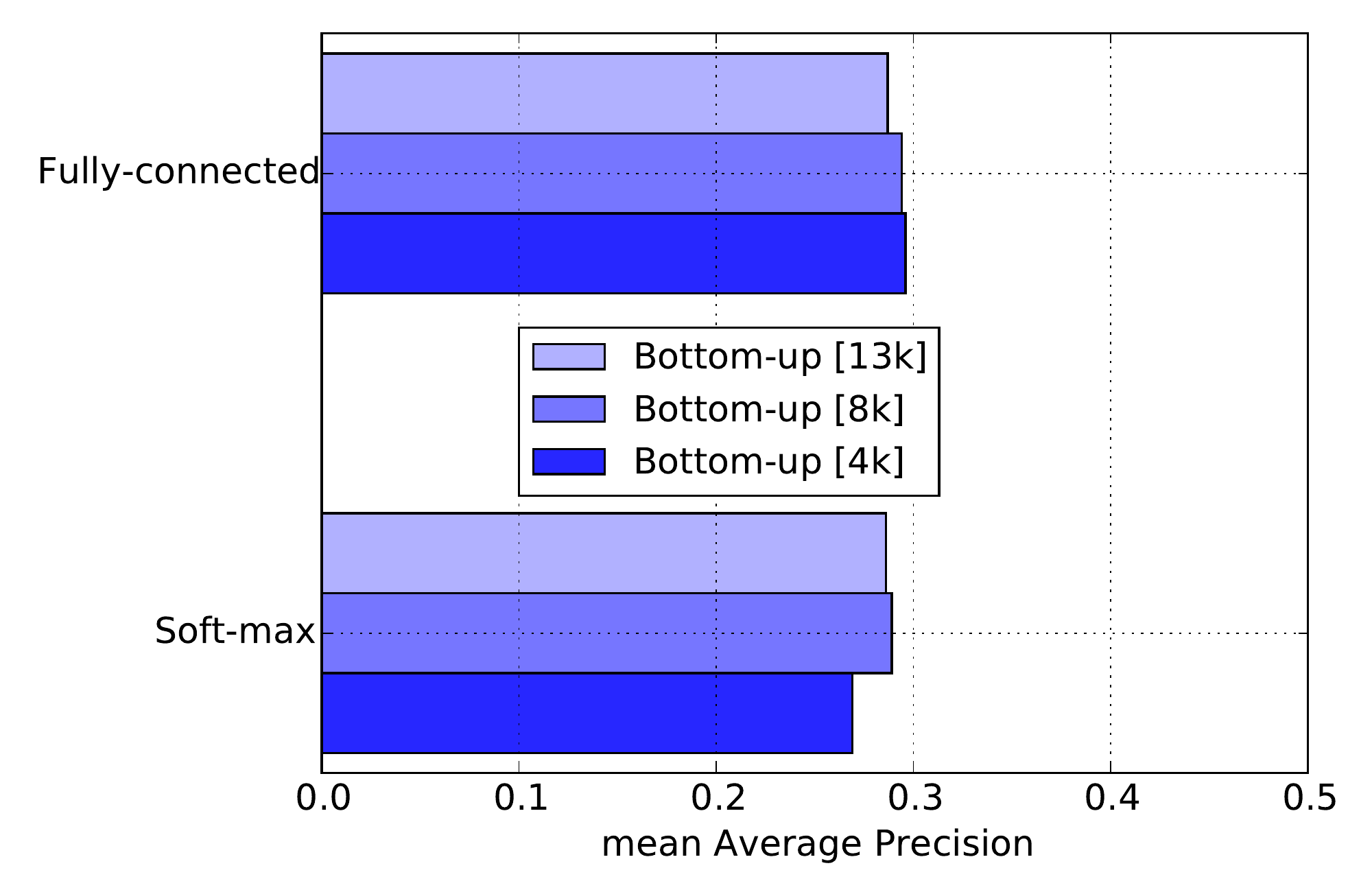}
\caption{TRECVID MED 2013 with 10 positives.}
\end{subfigure}
\caption{Mean Average Precision scores for the three bottom-up variants on TRECVID Multimedia Event Detection 2013. We observe that the more classes are maintained in the bottom-up reorganization, the better the performance using the soft-max (i.e. semantic) layer. The reserve happens for the fully-connected layer.}
\label{fig:exp1-bottomup}
\end{figure*}

\section{Experimental setup}

\subsection{Dataset}
\noindent
\textbf{TRECVID Multimedia Event Detection 2013.} The TRECVID Multimedia Event Detection 2013 dataset consists of roughly 27,000 test videos~\cite{over14}. The dataset contains annotations for 20 everyday events, including \emph{Birthday Party}, \emph{Making a sandwich}, \emph{Attempting a bike trick}, and \emph{Dog show}. The dataset has two different tasks, one where 10 positive videos are given for each event (10 Ex.), and one where 100 positive videos are given for each event (100 Ex.). For an event, a classifier is trained on the 10 or 100 positive videos and a background set of roughly 5,000 negative videos. The classifier is in turn used to rank the 27,000 test videos and its performance is evaluated using the (mean) Average Precision score on the ranked test videos.

\subsection{Implementation details}

\textbf{Deep convolutional networks.} We focus our evaluation on the recent GoogLeNet of Szegedy \etal~\cite{szegedy2014going}. The GoogLeNet is a deep convolutional neural network consisting of 22 layers. We also compare against the AlexNet of Krizhevsky \etal~\cite{krizhevsky2012imagenet}. The AlexNet consists of 5 convolutional layers and 3 fully-connected layers. To pre-train the deep networks, we utilize the open-source Caffe library~\cite{jia2014caffe} and the provided layer definitions and hyper-parameters for both networks.

\textbf{Feature extraction.} After pre-training, we extract features both at the fully connected layer and the soft-max layer. In AlexNet, we use the features from the second fully-connected layer, with a 4,096-dimensional frame representation. In GoogLeNet, we use the features at the pool5 layer, with a 1,024-dimensional frame representation. The dimensionality at the soft-max layer, which provides a probability score of each concept, for both networks is equal to the number of classes in the corresponding hierarchy.

\textbf{Pooling and Event Classification.} For event detection, we average the representations of the frames over each video unless stated otherwise, followed by $\ell_{1}$-normalization. We train an SVM classifier for each event separately with a $\chi_{2}$ kernel. We set the $C$ parameter to 100 in all our experiments.

\section{Experiments}
We consider four experiments. First, we evaluate the effect of different settings of the operations in our bottom-up reorganized pre-training. Second, we compare standard pre-training versus both the bottom-up and top-down reorganized pre-training. Third, we perform various fusions between deep representations and representation from other modalities. Fourth, we compare our results to the state-of-the-art on multimedia event detection.

\subsection{Bottom-up operation parameters}

\textbf{Experiment 1.} For the first experiment, we investigate the parameters for the bind and promote operations in our bottom-up reorganization, which have a significant influence on the amount of remaining classes. In total, we have trained three separate GoogLeNets~\cite{szegedy2014going} based on different parameters for the bind and promote operations:
\begin{itemize}
\item \textbf{Bottom-up [4k]:} Deep network pre-trained on 4,437 classes with $t_b=7,000$ and $t_p=1,250$.
\item \textbf{Bottom-up [8k]:} Deep network pre-trained on 8,201 classes with $t_b=7,000$ and $t_p=500$.
\item \textbf{Bottom-up [13k]:} Deep network pre-trained on 12,988 classes with $t_b=3,000$ and $t_p=200$.
\end{itemize}
For all the variants, we set the subsample threshold to $t_s=2,000$. An overview of mean Average Precision scores using the fully-connected and soft-max layers on TRECVID Multimedia Event Detection 2013 is shown in Figure~\ref{fig:exp1-bottomup}. We report the mean Average Precision scores both for the task with 10 positive videos and with 100 positive videos per event.

\textbf{Results.} From Figure~\ref{fig:exp1-bottomup}, we observe that the best scores using the fully-connected layer are achieved with the bottom-up [4k] variant. This result shows that the fully-connected layer translates best to events when merging more classes into a generic hierarchy. Interestingly, we observe the reverse pattern for the soft-max layer; the more classes are maintained the better the event detection performance. This result follows the work of Habibian et al.~\cite{habibian13}, which states that using more semantic classifiers is preferred over using better semantic classifiers. Here, we show that this observation translates to deep networks for event detection.

From this experiment, we conclude that the choice of the bottom-up reorganized variant depends on the desired deep network representation. The highest overall results are achieved by the features from the non-semantic fully-connected layer of the variant from 4,437 classes (0.446 mean Average Precision using 100 positives per event, 0.296 using 10 positives). However, the  variant from 12,988 classes performs best using the semantic features from the soft-max layer (0.441 using 100 positives, 0.286 using 10 positives).

\begin{figure*}[t]
\centering
\begin{subfigure}{0.49\textwidth}
\includegraphics[width=\textwidth]{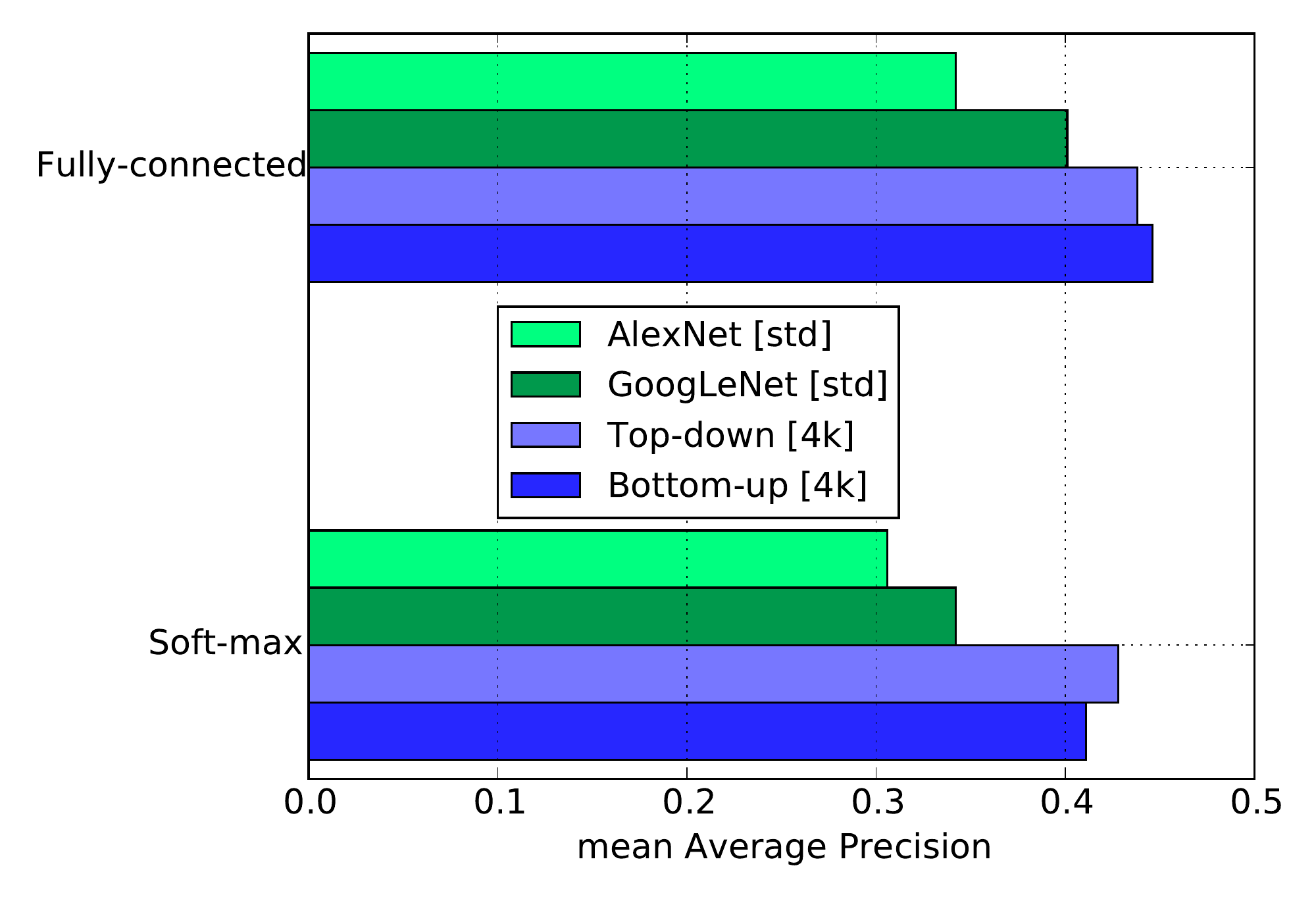}
\caption{TRECVID MED 2013 with 100 positives.}
\end{subfigure}
\begin{subfigure}{0.49\textwidth}
\includegraphics[width=\textwidth]{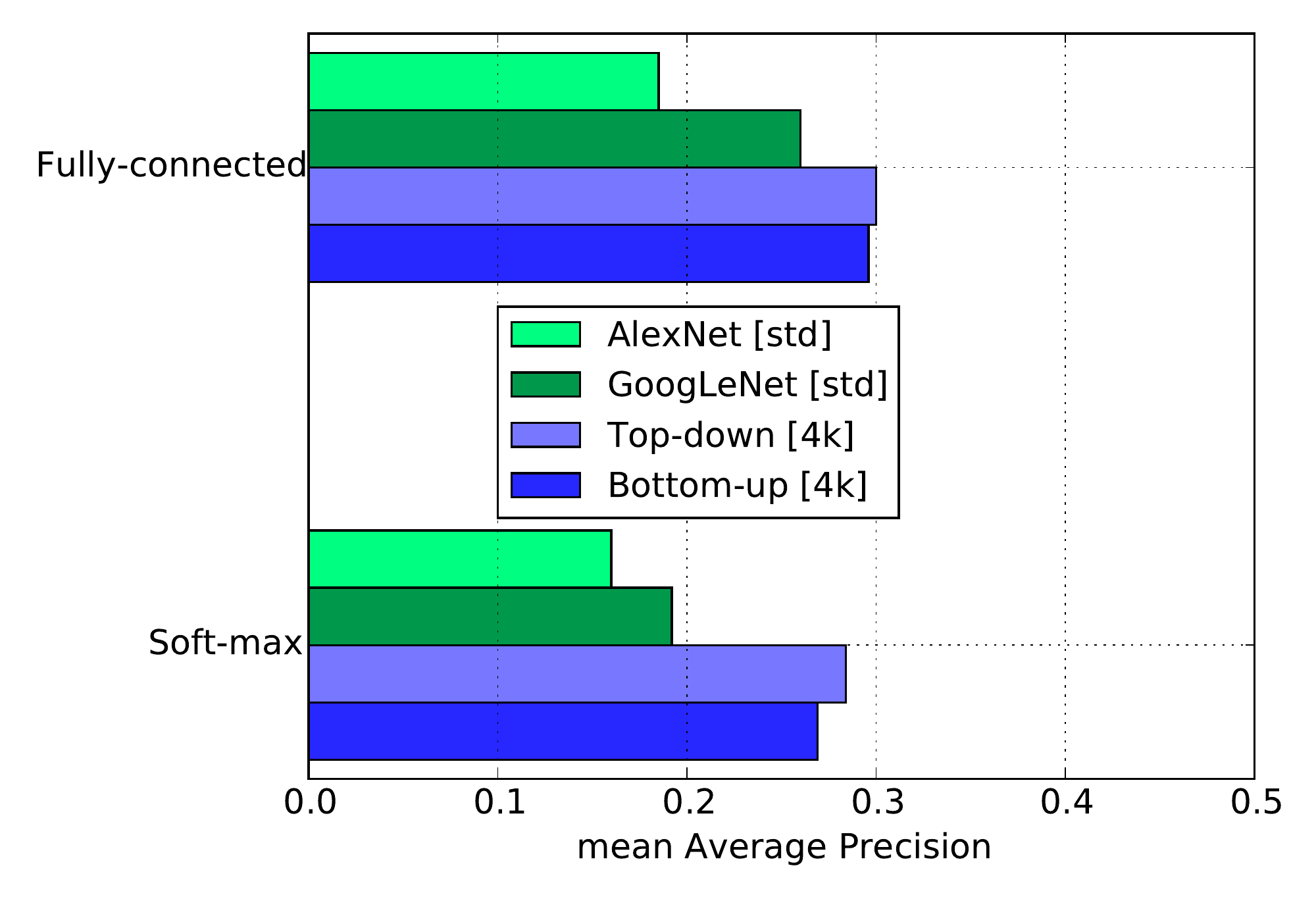}
\caption{TRECVID MED 2013 with 10 positives.}
\end{subfigure}
\caption{Mean Average Precision scores for our bottom-up and top-down reorganized pre-training (blue), compared to standard pre-training (green) on TRECVID MED 2013. Our approaches both clearly outperform standard pre-training, while being competitive and potentially complementary to each other.}
\label{fig:exp2-compare}
\end{figure*}

\subsection{Standard versus reorganized pre-training}
\textbf{Experiment 2.} For the second experiment, we compare our bottom-up and top-down reorganized pre-training against the conventional pre-training setup using the ImageNet 1,000 class subset~\cite{russakovsky2014imagenet}. For all networks, we report the Average Precision scores using both the fully-connected layer and the soft-max layer. For the bottom-up approach, we use the deep network pre-trained on 4,437 classes. For the top-down approach, we select the top 4,000 classes, with $t_t = 1,200$ for the threshold on the number of images required for each class. We compare our two approaches to two standard pre-trained deep networks:
\begin{itemize}
\item \textbf{AlexNet [std]:} AlexNet pre-trained on 1,000 ImageNet classes~\cite{krizhevsky2012imagenet}.
\item \textbf{GoogLeNet [std]:} GoogLeNet pre-trained on 1,000 ImageNet classes~\cite{szegedy2014going}.
\end{itemize}

\textbf{Results.} An overview of the comparison between standard and reorganized pre-training is shown in Figure~\ref{fig:exp2-compare}. We observe that the top-down and bottom-up reorganization approaches achieve comparable performance. While bottom-up performs slightly better using the fully-connected layer, top-down performs slightly better using the soft-max layer. We also note our reorganized pre-training approaches on GoogLeNet significantly outperform the standard pre-trained GoogLeNet. This holds especially for the soft-max layer, where the difference between standard pre-training and our top-down pre-training is 8.6\% and 9.2\% in absolute mean Average Precision for respectively the 100 and 10 positive video tasks. Lastly, we note that the difference in performance to the pre-trained AlexNet is even bigger. This result shows that GoogLeNet provides overall better visual representations, leading to improved event detection.

From this experiment, we conclude that our two approaches to reorganized ImageNet pre-training yield strong event detection results and significantly improve over standard pre-trained deep networks.

\subsection{Fusing representations and modalities}

\textbf{Experiment 3.} For the third experiment, we investigate the effect of feature fusion. Here, fusion is performed in a late fashion, by averaging the classifier scores of different classifiers. We investigate feature fusion in two aspects: \textit{i)} we investigate the effect of fusing different layers and video encodings from different pre-trained deep networks for event detection, \textit{ii)} we investigate the effect of fusing our deep visual representations with two other representations:
\begin{itemize}
\item \textbf{Audio modality:} MFCC features with first and second order derivatives, 30 dimensions for each of the three features, aggregated into a 46,080-dimensional video representation using Fisher Vectors with 256 clusters.
\item \textbf{Motion modality:} MBHx, MBHy, and HOG features computed along dense trajectories~\cite{wang13}, reduced to 128 dimensions using PCA, aggregated into a 65,536-dimensional video representation using Fisher Vectors with 256 clusters.
\end{itemize}

\textbf{Results for Fusing Networks.} 
In Table~\ref{tab:fusedeep}, we show an overview of fusion results using deep networks. Comparing index (1) to (3) and comparing index (2) to (4), we see that for both the bottom-up and top-down approach, it is beneficial to fuse the scores from the fully-connected and soft-max layers. This result is surprising, given that the layers come from the same network and it indicates that the layers contain different information useful for event detection. This result is furthermore interesting from a computational perspective, as the features from both layers can be extracted from a single pass through the same network. Hence, the improvement is obtained for free.

\begin{figure}[t]
\includegraphics[width=\linewidth]{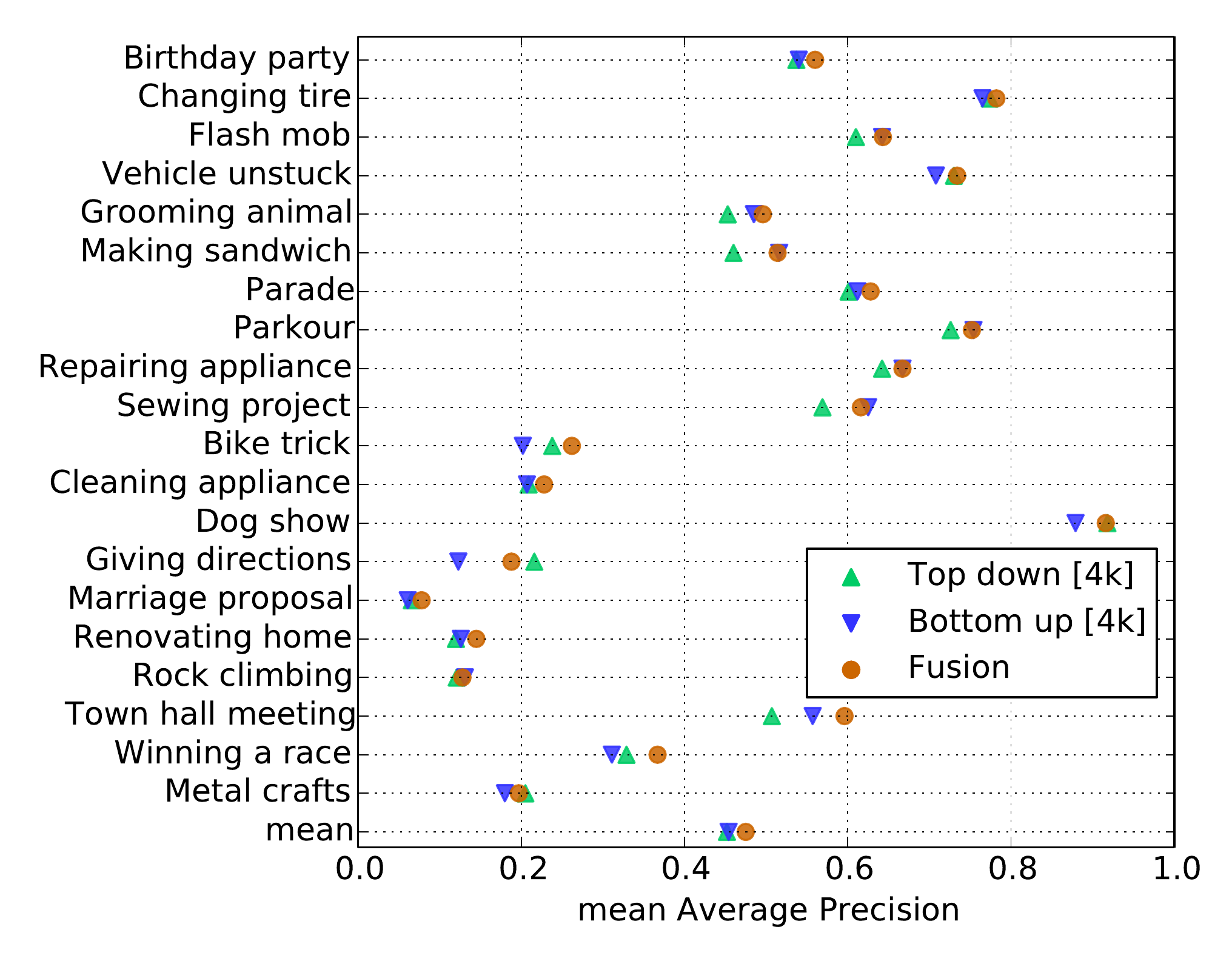}
\caption{The Average Precision scores per event for bottom-up, top-down, and their fusion. Note that the scores per event are different for the two approaches, resulting in complementary fusion results.}
\label{fig:deepfuseperevent}
\end{figure}

The fusion of (3) and (4), i.e. the fusion of the bottom-up and top-down reorganizations, also yields complementary results, with a mean Average Precision of 0.475 and 0.324 for respectively the 100 and 10 positive video tasks of the TRECVID Multimedia Event Detection 2013 dataset. This result clearly shows that pre-training deep networks on different hierarchies results in different and complementary representations. Figure~\ref{fig:deepfuseperevent} shows that, although the mean Average Precision of the individual approaches is similar, the scores per event vary notably (on average 2.7\% per event), resulting in improved performance upon fusion.

\begin{table}[t]
\centering
\begin{tabular}{l r r}
\toprule
\multicolumn{3}{c}{\textbf{TRECVID MED 2013}}\\
Method & 100 Ex. & 10 Ex.\\
\midrule
\textbf{Averaging} & &\\
(1) Bottom-up (fc) & 0.446 & 0.296\\
(2) Top-down (fc) & 0.438 & 0.300\\
(3) Bottom-up (fc + soft-max) & 0.452 & 0.305\\
(4) Top-down (fc + soft-max) & 0.454 & 0.317\\
(3) + (4) & 0.475 & 0.324\\
\midrule
\textbf{VLAD} & &\\
Bottom-up (fc + soft-max) & 0.465 & 0.339\\
\bottomrule
\end{tabular}
\caption{Mean Average Precision scores for fusions of different layers, networks, and encodings within deep representations, which all yield complementary results.}
\label{tab:fusedeep}
\end{table}

Multiple recent works have investigated complex and high-dimensional video representations from deep frame representations beyond frame averaging~\cite{nagel2015event,xu2014discriminative}. Here, we similarly investigate such representations using the frame features with reorganized pre-training. We have employed both VLAD and Fisher Vector encoding and report the results for VLAD in Table~\ref{tab:fusedeep}, as that yielded the highest scores. For the VLAD encoding, we create a codebook from 10 clusters per event, resulting in a 10,240-dimensional feature vector using the fully-connected layer and a 44,370-dimensional feature vector using the soft-max layer. The results using the bottom-up reorganization show that a VLAD encoding improves over averaging, especially for the 10 positive videos per event task (0.339 mAP versus 0.305 for averaging).

We conclude from this fusion experiment that combining information from different pre-trained deep networks and even different layers from the same deep network improves the Average Precision scores. Furthermore, performing a VLAD encoding instead of averaging frames results in a boost for individual networks, especially for the 10 positive videos task.

\textbf{Results for Fusing Modalities.} 
In Table~\ref{tab:fuseother}, we show the results of the deep networks with the audio and motion modalities. The Table clearly states that individually, the event detection scores using our deep networks improve over the motion and audio scores. Upon combining the modalities, we observe a jump in performance. This result shows the complementary natures of the different modalities: individually the motion and audio features are clearly outperformed, but they contain information not captured in deep networks which result in improved fusion results. This is naturally due to the nature of deep convolutional neural networks, which focus on spatially visual information and exclude temporal and audio information.

We have furthermore attempted to fuse the VLAD encoding of Table~\ref{tab:fusedeep} with the motion and audio features, but this did not result in improved performance. Since the VLAD encoding requires more computational effort and has a higher storage requirement, we have opted to focus on averaging.

\subsection{Comparison to the state-of-the-art}
\textbf{Experiment 4.} For the fourth experiment, we compare our results to the current state-of-the-art on multimedia event detection. We perform a comparison on both the TRECVID MED 2013 test set and the TRECVID MED 2015 benchmark.

\textbf{Results on the TRECVID MED 2013 Test set.} The comparison to the state-of-the-art on the TRECVID Multimedia Event Detection 2013 dataset is shown in Table~\ref{tab:sota2013}. As the Table shows, we outperform the current state-of-the-art on both the 100 and 10 positive videos per event task using deep networks only. Upon a fusion with motion and audio features, we improve further over related work.

\begin{table}[t]
\centering
\begin{tabular}{l r r}
\toprule
\multicolumn{3}{c}{\textbf{TRECVID MED 2013}}\\
Method & 100 Ex. & 10 Ex.\\
\midrule
\textbf{Individual} & &\\
(1) Audio (MFCC) & 0.114 & 0.053\\
(2) Motion (MBH) & 0.341 & 0.192\\
(3) Visual (ours, avg) & 0.475 & 0.324\\
\midrule
\textbf{Fusion} & &\\
(2) + (3) & 0.504 & 0.345\\
(1) + (2) + (3) & 0.526 & 0.348\\
\bottomrule
\end{tabular}
\caption{Mean Average Precision scores for fusing our deep representations with other modalities. We outperform motion and audio features, while the fusion leads to further improvement.}
\label{tab:fuseother}
\end{table}

\begin{table}[t]
\centering
\begin{tabular}{l r r}
\toprule
\multicolumn{3}{c}{\textbf{TRECVID MED 2013}}\\
Method & 100 Ex & 10 Ex\\
\midrule
Habibian et al. \cite{habibian14} 		& - 	& 0.196\\
Sun et al. (visual) \cite{sun142} 	    & 0.350 & -\\
Nagel et al. \cite{nagel2015event} 		& 0.386 & 0.218\\
Sun et al. (fusion) \cite{sun142}        & 0.425 & -\\
Xu et al. \cite{xu2014discriminative} 	& 0.446 & 0.298\\
Chang et al. \cite{chang2015searching} 	& - 	& 0.310\\
\midrule
Ours, deep network 						& 0.475 & 0.324\\
Ours, multimodal fusion 				& 0.526 & 0.348\\
\bottomrule
\end{tabular}
\caption{Comparison to other works on TRECVID MED 2013 test set for both our best deep network results and our fusion results. We yield better results for both the 100 and 10 positive videos per event task.}
\label{tab:sota2013}
\end{table}

\begin{figure*}[t]
\centering
\begin{subfigure}{0.32\textwidth}
\includegraphics[width=\textwidth]{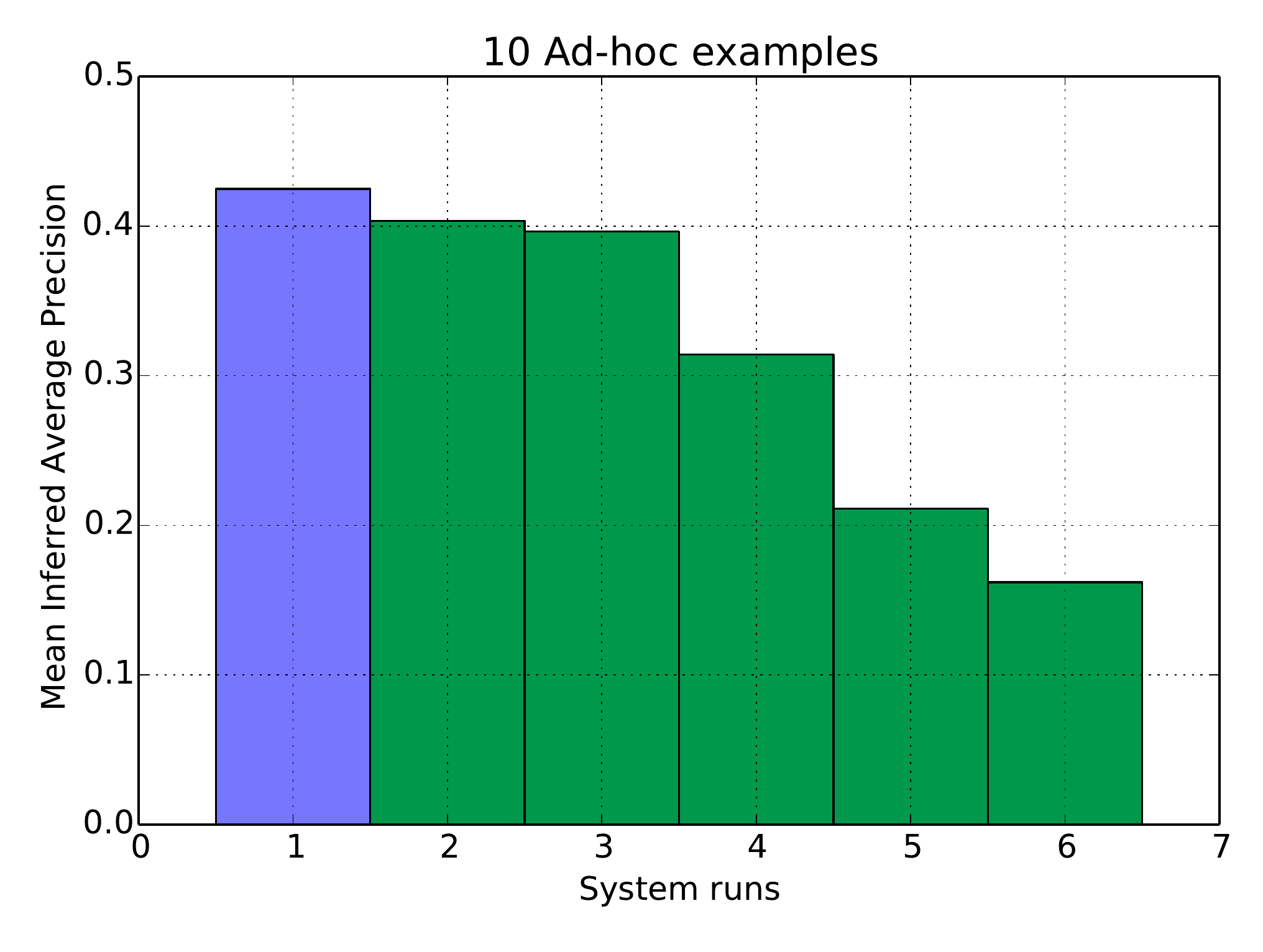}
\end{subfigure}
\begin{subfigure}{0.32\textwidth}
\includegraphics[width=\textwidth]{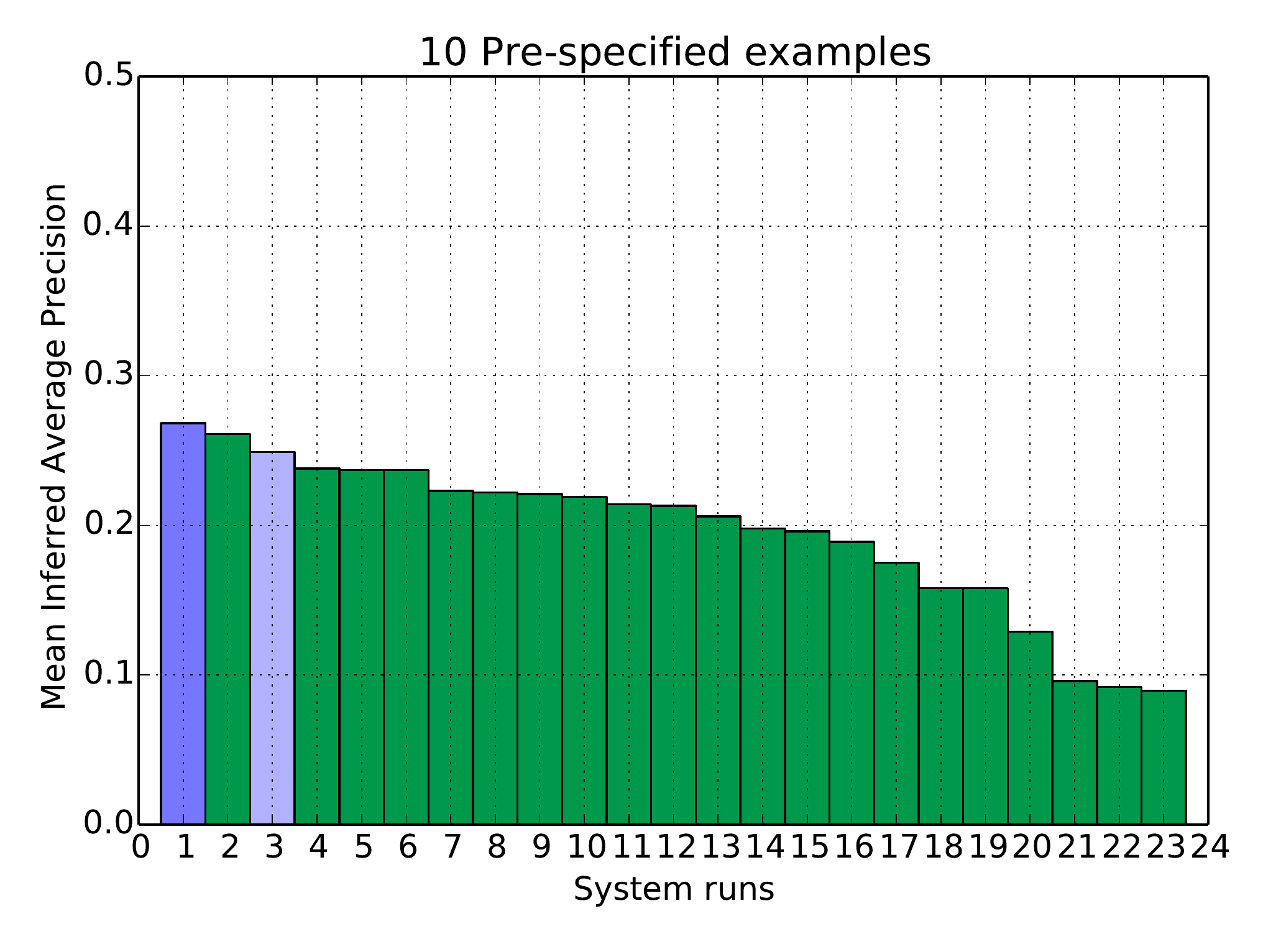}
\end{subfigure}
\begin{subfigure}{0.32\textwidth}
\includegraphics[width=\textwidth]{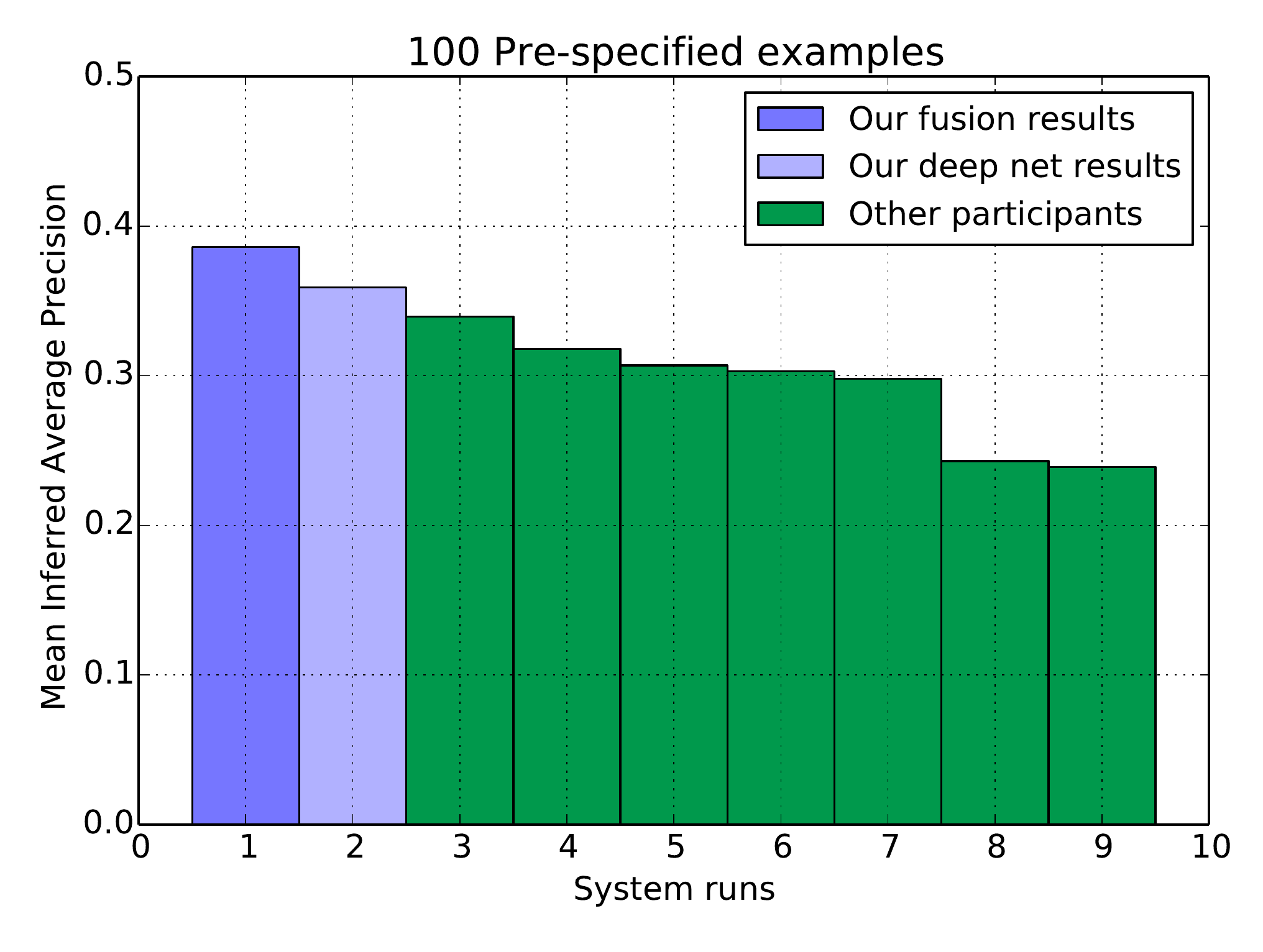}
\end{subfigure}
\caption{Comparison between our results (blue) and the results of all other participants in the TRECVID Multimedia Event Detection benchmark 2015. Our deep networks and their fusion with motion and audio information are the top contenders for all tasks.}
\label{fig:sota2015}
\end{figure*}

\textbf{Results on the TRECVID MED 2015 Benchmark.} We furthermore compare our results achieved on the latest TRECVID 2015 benchmark for Multimedia Event Detection. This benchmark is similar in nature to the 2013 dataset in training and evaluation. However, the 2015 dataset contains 20 new events. Furthermore, the benchmark comparison is performed in a large-scale setup, with a test set of about 200,000 videos. In Figure~\ref{fig:sota2015}, we show the inferred mean Average Precision scores for our entries and the entries of the other participants. We report results both for the pre-specified (where the events and video labels are given well before the benchmark deadline) and ad-hoc (where the events and video labels are given shortly before the benchmark deadline) tasks. The Figure paints a similar picture to the results on the 2013 dataset; we outperform the current state-of-the-art by fusing deep representations with motion and audio modalities, while our deep representations only are already among the top contenders.

\section{Conclusions}
In this work, we leverage the complete ImageNet dataset for pre-training deep convolutional neural networks for video event detection, rather than the prescribed 1,000 ImageNet subset. We propose two contrasting and complementary approaches to reorganize the ImageNet hierarchy. The bottom-up approach aims to merge classes from the deepest parts of hierarchy upwards, while the top-down approach aims to select rich generic classes starting from the top of the hierarchy. The new hierarchies are in turn used as input to pre-train deep networks and are employed for frame representation in video event detection. Experimental evaluation performed on the challenging TRECVID MED 2013 dataset shows that deep networks trained on our hierarchies \textit{i)} outperform standard pre-trained networks, \textit{ii)} are complementary, \textit{iii)} maintain the benefits of fusion with other modalities, and \textit{iv)} reach state-of-the-art result. The pre-trained models are available online at
\url{http://tinyurl.com/imagenetshuffle}
and can be used directly to extract state-of-the-art video representations using the Caffe library.

\section*{Acknowledgements}
This research is supported by the STW STORY project.
\vspace{-0.2cm}

\bibliographystyle{abbrv}
\bibliography{the-imagenet-shuffle-arxiv}

\end{document}